\documentclass{article}

\usepackage{PRIMEarxiv}

\usepackage[utf8]{inputenc} % allow utf-8 input
\usepackage[T1]{fontenc}    % use 8-bit T1 fonts
\usepackage{hyperref}       % hyperlinks
\usepackage{url}            % simple URL typesetting
\usepackage{booktabs}       % professional-quality tables
\usepackage{amsfonts}       % blackboard math symbols
\usepackage{nicefrac}       % compact symbols for 1/2, etc.
\usepackage{microtype}      % microtypography
\usepackage{lipsum}
\usepackage{fancyhdr}       % header
\usepackage{graphicx}       % graphics
\usepackage{amsmath}
\usepackage{amsfonts}
\usepackage{multicol}
\usepackage{multirow}
\usepackage{mwe}
\usepackage{pdfpages}
\usepackage{float}
\usepackage{amssymb}
\usepackage{algorithm}  
\usepackage{algorithmic}  
\usepackage{subfigure}
\usepackage{color}
\usepackage{xcolor}
\graphicspath{{media/}}     % organize your images and other figures under media/ folder

\author{
    Xueru Wen \\
    College of Computer Science and Technology \\
    Jilin University \\
    Changchun\\
    \texttt{wenxr2119@mails.jlu.edu.cn} \\
\And
    Changjiang Zhou \\
    College of Computer Science and Technology \\
    Jilin University \\
    Changchun\\
\And
    Haotian Tang \\
    College of Computer Science and Technology \\
    Jilin University \\
    Changchun\\
\And
    Luguang Liang \\
    College of Computer Science and Technology \\
    Jilin University \\
    Changchun\\
\And
    Yu Jiang \\
    Key Laboratory of Symbolic Computation and Knowledge Engineering of Ministry of Education \\
    Jilin University \\
    \texttt{jiangyu2011@jlu.edu.cn} \\
\And
    Hong Qi \\
    Key Laboratory of Symbolic Computation and Knowledge Engineering of Ministry of Education \\
    Jilin University \\
}

\title{Type-supervised sequence labeling based on the heterogeneous star graph for named entity recognition}

\begin{document}
\maketitle

\begin{abstract}
Named entity recognition is a fundamental task in natural language processing, identifying the span and category of entities in unstructured texts.
The traditional sequence labeling methodology ignores the nested entities, i.e. entities included in other entity mentions.
Many approaches attempt to address this scenario, most of which rely on complex structures or have high computation complexity.
The representation learning of the heterogeneous star graph containing text nodes and type nodes is investigated in this paper.
In addition, we revise the graph attention mechanism into a hybrid form to address its unreasonableness in specific topologies.
The model performs the type-supervised sequence labeling after updating nodes in the graph.
The annotation scheme is an extension of the single-layer sequence labeling and is able to cope with the vast majority of nested entities.
Extensive experiments on public NER datasets reveal the effectiveness of our model in extracting both flat and nested entities.
The method achieved state-of-the-art performance on both flat and nested datasets. 
The significant improvement in accuracy reflects the superiority of the multi-layer labeling strategy.
\end{abstract}

\keywords{Named Entity Recognition \and Sequence Labeling \and Heterogeneous Graph}

\twocolumn

\section{Introduction}
Named Entity Recognition is an essential task in natural language processing that aims to recognize the boundaries and types of entities with specific meanings in the text, including names of people, places, institutions, etc.
The Named Entity Recognition task is not only a vital tool for information extraction, but also a crucial component in many downstream tasks, such as text understanding \cite{2019Attending}.

\begin{figure}[H]
	\centering
	\includegraphics[width=0.48\textwidth]{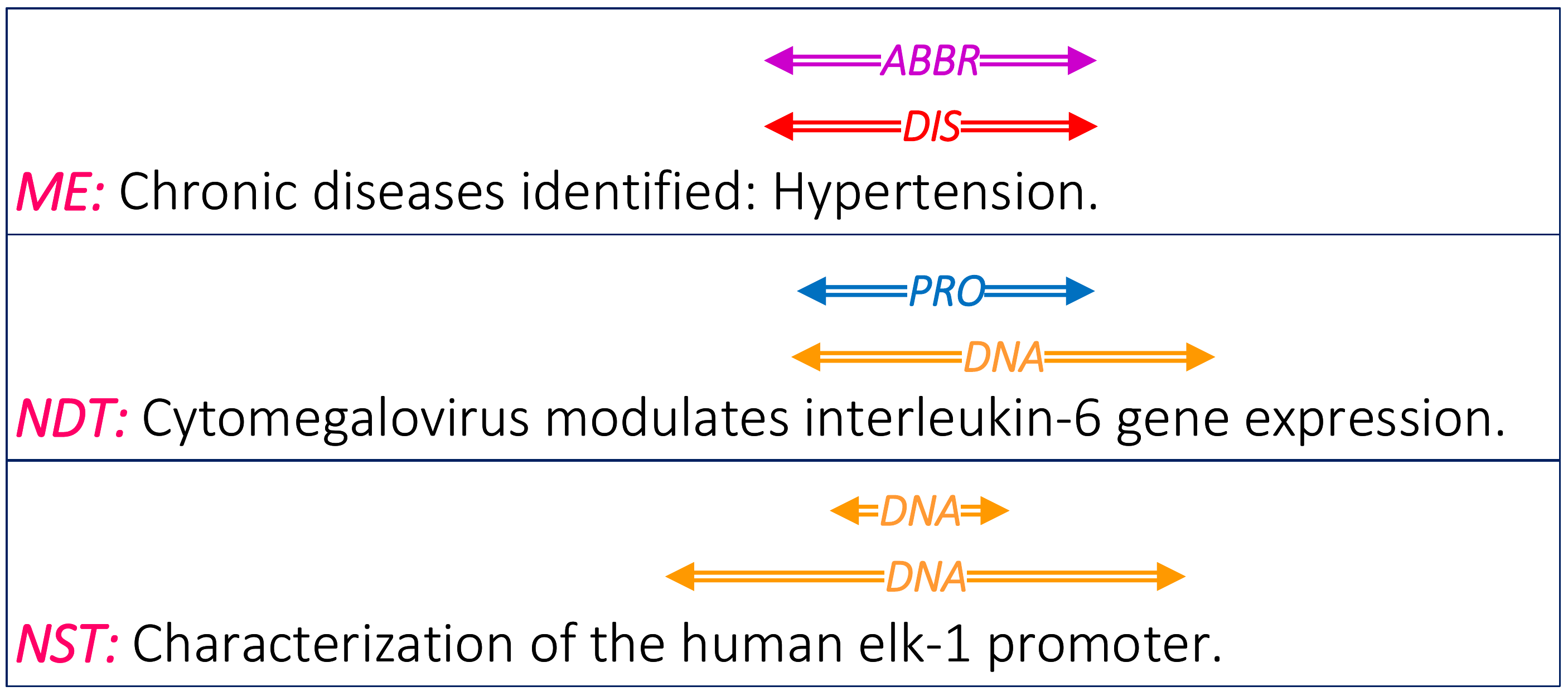}
	\caption{Example of entity nesting from GENIA \cite{kim2003genia} and Chilean Waiting List \cite{rojas-etal-2022-simple}. The colored arrows indicate the category and span of the entities. The bolded black abbreviations denote the type of entity nesting.}
	\label{nest_example}
\end{figure}

Named entity recognition is usually modeled as a sequence labeling problem and can be efficiently solved by an RNN-based approach \cite{2015Bidirectional}.
The sequence labeling modeling approach simplifies the problem based on the assumption that entities never nested with each other.
However, entities may be overlapping or deeply nested in real-world language environments, as in Figure \ref{nest_example}.
More and more studies are exploring modified models to deal with this more complex situation.

Some works like \cite{TakashiShibuya2020NestedNE} employ a layered model to handle entities nesting, which iteratively utilizes the result of the previous layer to be further annotated until reaches the maximum number of iterations or generate no more new entities.
Nevertheless, these models suffer from the problem of interlayer disarrangement, that is, the model may output a nested entity from a wrong layer and pass the error to the subsequent iterations.
The main reason for this phenomenon is that the target layer to generate the nested entity is determined by its nesting levels rather than its semantics or structure.

Some other work like \cite{shen2021locate,XiaoyaLi2020AUM} identifies nested entities by enumerating entity proposals.
Although these methods are theoretically perfect, they still confront difficulties in model training, high complexity, and negative samples.
These obstacles stem from the fact that the enumeration approach does not take into account the a priori structural nature of nested entities.

In recent years, graph neural networks have received a lot of attention.
Most early graph neural networks like \cite{VictorGarciaSatorras2018FewshotLW} are homogeneous graphs.
But the graphs encountered in practical applications are generally heterogeneous graphs with nodes and edges of multiple types.
An increasing number of studies are dedicated to applying graph models in NLP tasks.
Among them, \cite{XIE2020105548} introduces a heterogeneous document entity graph for multi-hop reading comprehension containing information at multiple granularities.
And \cite{wang2020heterogeneous} proposes a neural network for summary extraction based on heterogeneous graphs with semantic nodes of different granularity levels, including sentences.

In this paper, we design a multi-layer decoder for the NER task.
To address the interlayer disarrangement, the model groups entities directly according to their categories, instead of grouping entities based on the nesting depth.
Each layer individually recognizes entities of the same category.
This method extends the traditional sequence labeling method and eases the problem of nested entities to a certain extent.
Meanwhile, this annotation method can recognize multi-label entities overlooked by most models targeting the nested NER task.
This nesting scenario is first mentioned in \cite{alex-etal-2007-recognising}, and is very common in some datasets like \cite{baez-etal-2020-chilean}.
In addition, to deal with the case of the nested entities of the same type, this paper designs an extended labeling and decoding scheme that further recognize nested entities in a single recognition layer.
The proposed type-supervised sequence labeling model can naturally combine with a heterogeneous graph.
For this purpose, we propose a heterogeneous star graph model.

In summary, the contributions of our work are as follows:
\begin{itemize}
    \item To the best of our knowledge, we are the first to apply the heterogeneous graph in the NER task. The proposed graph network efficiently learns the representation of nodes, which can be smoothly incorporated with the type-supervised sequence labeling method. Our model achieved state-of-the-art performance on flat and nested datasets \footnote{Access the code at \href{https://github.com/Rosenberg37/GraphNER}{https://github.com/Rosenberg37/GraphNER}}.
    \item We design a stacked star graph topology with type nodes as the center and text nodes as the planetary nodes. It greatly facilitates the exchange of local and global information and implicitly represents location information. This graph structure also significantly reduces the computational complexity to $O(tn)$ from the $O(n^2)$ of general attention mechanisms.
    \item Our graph attention mechanism is proposed for addressing the specific scenarios in which traditional graph attention mechanisms fail. The favorable properties of our attention mechanism can naturally express the edge orientation.
    \item The proposed type-supervised labeling method and the corresponding decoding algorithm not only can recognize vast majority of nested entities but also cope with the cases neglected by most nested entity recognition models.
\end{itemize}

\section{Related Work}

\subsection{Named Entity Recognition}
In recent years, named entity recognition models based on deep learning have been the main direction of relevant research.
Deep learning approaches enhance the model's ability of the feature representation and data fitting by automatically mining hidden features without human intervention.
Models like \cite{YueZhang2018ChineseNU} based on recurrent neural networks and conditional random fields have become the dominant baseline models.

Transformer proposed in \cite{2017Attention} comprehensively employs the attention mechanism to construct an encoder-decoder framework and shows satisfactory performance in many NLP tasks.
Star Transformer presented in \cite{guo2019star} discards the fully connected structure in the original construction and achieves low computational complexity and implicit representation of the position information.
It's applied to the downstream Chinese NER task in \cite{chen2021enhancing} and obtains outstanding results. 
In our work, we extend the star-connection topology to construct a heterogeneous graph.

Since the classical NER has been comparatively sophisticated, nested entities recognition has gradually become the research hotspot.
Some works like \cite{ju-etal-2018-neural} deal with nested entities in layered models.
They predict entities in an inside-to-outside order by dynamically stacked LSTM-CRF layers.
Nevertheless, layered models are burdened with error propagation caused by identifying entities at the inaccurate layer.
Region-based methods such as \cite{sohrab-miwa-2018-deep} identify nested entities by enumerating all possible spans in text and classifying them.
However, these methods suffer from high computational complexity and difficulties in model training.
In this paper, we propose a type-supervised sequence labeling scheme to resolve these problems.

\subsection{Graph Neural Network}
Graph Neural Networks like \cite{zhou2021graph} can capture dependencies through passing messages between nodes on the graph.
Due to the needs of real-world scenarios, the design and application of heterogeneous graph neural networks has attracted extensive interest.
\cite{zhao2021representation} proposes a graph neural network based on heterogeneous graph iterations to resolve the problem of relation extraction in the presence of overlap.
\cite{gui2019lexicon} combine the lexicon with GNN and apply it in Chinese NER.

The employment of graph neural networks in NLP tasks has been widely explored.
In this paper, the types of entities are modeled as nodes on the graph to construct the heterogeneous graph.
We further utilize them in the subsequent sequence labeling.
In particular, the specifically designed topology structure of the graph allows for a reduction in computational complexity and an improvement in the interaction between global and local messages.

\section{Task Definition}
The goal of the named entity recognition task is to identify all possible entities in the input sentence.
For a given input sentence $\mathcal{S}=[w_1,w_2,... ,w_L]$, where $L$ is the length of the sentence.
The entity $x$ is defined as a triple $(s,e,t)$, where $s,e\in [1,L]$ denote the start and end indices of the entity and $t$ stands for the predefined entity category. 
With the definition, NER task can be expressed formally recognize the entity set $\mathcal{X}=\{x_1,x_2,...,x_M\}$ existing in the sentence $\mathcal{S}$.
We develop the definition of nested entities in \cite{rojas-etal-2022-simple} as follows:

\paragraph{\textbf{Multi-label Entities(ME)}}
For two entities $x_1$ and $x_2$, we call them multi-label entities if $(s_1=s_2)\wedge(e_1=e_2)\wedge(t_1 \ne t_2)$, as in Figure \ref{nest_example}.

\paragraph{\textbf{Nested Entities of Same Type(NST)}}
For two entities $x_1$ and $x_2$, we call them nested entities of same type if $(e_1 \ge e_2 \ge s_2 \ge s_1)\wedge(t_1=t_2)$, as in Figure \ref{nest_example}.
In particular, if $(e_1 = e_2 = s_2 = s_1)\wedge(t_1 = t_2)$, then they are just one entity.

\paragraph{\textbf{Nested entities of Different Type(NDT)}}
For two entities $x_1$ and $x_2$, if $(s_1 \ge s_2 \ge e_2 \ge e_1) \wedge(t_1 \ne t_2)$, we call them nested entities of different type, as in Figure \ref{nest_example}.
However, if $(s_1=s_2)\wedge(e_1=e_2)$, it's actually \textbf{ME}.

\paragraph{\textbf{Overlapping Entities of Same Type(OST)}}
For the case $(e_1 > e_2 \ge s_1> s_2)\wedge(t_1=t_2)$, we call it overlapping entity of same type, which is not a case addressed in this paper.

\paragraph{\textbf{Overlapping Entities of Different Type(ODT)}}
For the case $(e_1> e_2 \ge s_2 > s_1)\wedge(t_1\ne t_2)$, we call it overlapping entities of different type. 
Although our model does not target this scenario, it is implicitly solved as the decoding procedure is separated between different entity types.

In this paper, two entities $x_1$ and $x_2$ are considered to be nested entities only when they are \textbf{ME}, \textbf{NST} or \textbf{NDT}.
We model the NER task as a type-supervised sequence labeling task and perform it with the fusion of type nodes and text nodes generated by the heterogeneous graph neural network.

\begin{figure*}[t]
    \includegraphics[width=\textwidth]{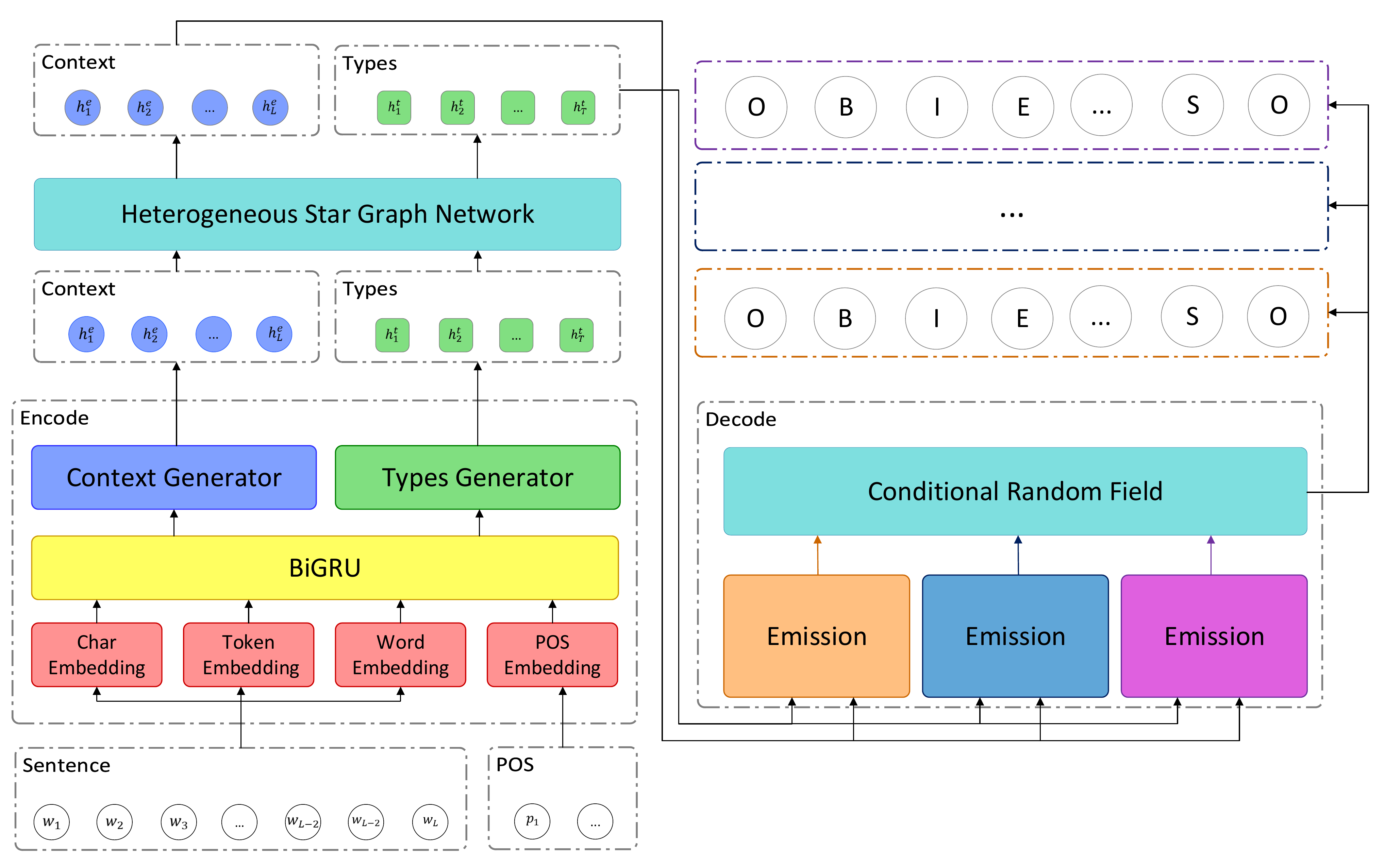}
    \caption{Overall architecture. In the figure and below, text nodes are represented using blue circles and type nodes are represented using green circles. Different colors in the Emission module and BIOES annotations indicate the recognition of corresponding classes of entities.}
    \label{Framework}
\end{figure*} 

\section{Methodology}
This section is going to detail our model.
The general framework is shown in Figure \ref{Framework} and consists of three main parts:

\begin{itemize}
    \item{\textbf{Node Representation}}
    Given the input sentence, the recurrent neural network is used to fuse characters, tokens, words, and part-of-speech annotation embeddings to produce the ultimate context presentation. The initial representation of the text nodes and type nodes are then generated from the context representation by linear transformation and pooling operation.
    \item{\textbf{Heterogeneous Graph}}
    The nodes update with the iteration of the star heterogeneous attention graph network. In this paper, we alter the concatenate-based graph attention mechanisms and take edge direction into consideration.
    \item{\textbf{Entity Extraction}}
    After getting the representation of each node, the text nodes are combined with the type nodes to produce the text representation under various types. To predict entity collection in the input sentences, we deploy the conditional random field to do the BIOES sequence labeling on each text representation. The union of each predicted entity set will be the ultimate collection of predicted entities.
\end{itemize}

\subsection{Node Representation}
The initialization of each node representation is required before the iteration of the graph neural network.
The heterogeneous graph in our paper consists of two kinds of nodes: type nodes and text nodes.
The following describes how to initialize each node's representation.
\subsubsection{Hybrid Embedding}
Before initializing the nodes, it is necessary to create the hidden representation of the context.
We use a multi-granularity hybrid embedding model to produce the context representation.

\paragraph{\textbf{Character}}
The embedded representation of characters can be formalized as follows.
\begin{equation}
[h^c_1,h^c_2,... ,h^c_D]=E_c([c_1,c_2,...,c_D])
\end{equation}
where $c_i$ is the one-hot code of the characters forming the word, $D$ is the number of characters constituting the word and $h^c_i$ is the embedding corresponding to $c_i$.
The characters' representations are then combined using recurrent neural networks and average pooling operation as follows:
\begin{equation}
h^C_i=\text{AvgPool}(\text{BiGRU}([h^c_1,h^c_2,... ,h^c_D]))
\end{equation}
where $\text{GRU}$ \cite{cho-etal-2014-learning} is the gated recurrent unit  and $h^C_i\in \mathbb{R}^{d_C}$ is the character-level hidden presentation for $w_i$.

\paragraph{\textbf{Token}}
The token-level presentation is generated by the pre-trained language model BERT \cite{devlin2019bert} which uses the Wordpiece partitioning \cite{wu2016google} to convert the tokens into subtokens.
The subtokens' representations are average pooled to produce the contextual representation of each token $w_i$. 
\begin{equation}
h^K_i=\text{AvgPool}([h^s_1,h^s_2,... ,h^s_O])
\end{equation}
where $h_i$ is the contextual embedding of the subtokens outputted from the language model, $O$ is the number of subtokens forming the word, and $h^K_i\in \mathbb{R}^{d_K}$ is the token-level embedding of $w_i$.

\paragraph{\textbf{Word}}
The embedding representation of a word can be formalized as follows:
\begin{equation}
[h^W_1,h^W_2,... ,h^W_L]=E_w([w_1,w_2,...,w_L])
\end{equation}
where $w_i$ denotes the one-hot code of the word, and $h^W_i\in \mathbb{R}^{d_W}$ is the word-level embedding. 
We exploit pre-trained word vectors in our work.

\paragraph{\textbf{Part-of-speech}}
We embedded representation of part-of-speech annotation as follows:
\begin{equation}
[h^P_1,h^P_2,... ,h^P_L]=E_p([p_1,p_2,...,p_L])
\end{equation}
where $p_i$ is the one-hot code of the part-of-speech tag and $h^P_i\in \mathbb{R}^{d_P}$ is the POS embedding.

The above embeddings are concatenated together:
\begin{equation}
    h^{\prime}_i = [h^K_i;h^C_i;h^W_i;h^P_i] 
\end{equation}
And then fused by the \text{BiGRU} network:
\begin{equation}
    \begin{aligned}
        H^{\prime} &= [h^{\prime}_1,h^{\prime}_2,...,h^{\prime}_L]\\
        H^A &= \text{BiGRU}(H^{\prime})\\
            &= [h^A_1,h^A_2,...,h^A_L]\\
    \end{aligned}
\end{equation}
Where $H^A$ denotes the ultimate text representation.

\subsubsection{Text Nodes}
Text nodes integrate the context information and are generated by a linear transformation of the context representation as follows:
\begin{equation}
    \begin{aligned}
        H^e &= W_e[h^A_1,h^A_2,... ,h^A_L]+b_e\\
            &= [h^e_1,h^e_2,... ,h^e_L]\\
    \end{aligned}
\end{equation}
where $h^e_i \in \mathbb{R}^{d_E}$ is the $i$-th text node's hidden state.

\subsubsection{Type Nodes}
Each type node corresponds to a specific predefined entity category.
Denote $H^a=[h^a_1,h^a_2,... ,h^a_L]$, the type node representation corresponding to each type $t$ can be derived by pooling the projected contextual representation:
\begin{equation}
    \begin{aligned}
        H^{t} &= W_{t}H^a+b_{t}\\
            h^{t} &= \text{MaxPool}(H^{t})    
    \end{aligned}
\end{equation}

\subsection{Heterogeneous Graph}
After determining the initial state of each node, the heterogeneous graph $\mathcal{G}=(\mathcal{V},\mathcal{E})$ can be built, and nodes are then iteratively updated.

$\mathcal{V}$ stands for the set of nodes and each $v \in \mathcal{V}$ belongs to a certain node type, denoted as $\phi(v) \in \mathcal{T}$, where $\mathcal{T}$ is the set of node types.
And for the heterogeneous graphs in our work, $\mathcal{T}=\{e,y\}$, which denote the node types of text and entity category, respectively.

$\mathcal{E}$ indicate the set of edges, and each $e\in \mathcal{E}$ can be written as a triple $(i,j,r)$, where $r\in \mathcal{R}$.
$\mathcal{R}$ is the set of edge types, and in this paper there are $\mathcal{R}=\{r,\tilde{r},\hat{r}\}$.
We denote the positive direction of the edge as $r$ and inverse direction of the edge as $\tilde{r}$, i.e., if there is $(i, j, r)$, then there is $(j, i, \tilde{r})$.
And $\hat{r}$ denotes self-connected edges, i.e., if there is $i=j$ in $(i, j, r)$ then there is $(i, j, \hat{r})$.

For a particular node $i$, the set of its $k$-order neighbors is defined recursively as $\mathcal{N}_i^k=\mathcal{N}_i^{k-1} \cup \{j|\exists i\in \mathcal{N}_i^{k-1} \wedge (i,j,r)\in \mathcal{E}\}$.
In particular, there is $\mathcal{N}_i^0=\{i\}$, i.e., the $0$-order set of neighbors is the single point set consisting of its own.
In fact, the set of $k$-order neighbors includes all nodes that can be reached by up to $k$-hops from node $i$.
Furthermore, we define the $k$-order punctured set of neighbors $\overline{\mathcal{N}_i^k} = \mathcal{N}_i^k \setminus \mathcal{N}_i^0$, i.e., the set of $k$-order neighbors without the node itself.

\subsubsection{Topology}
\begin{figure*}[t]
    \includegraphics[width=\textwidth]{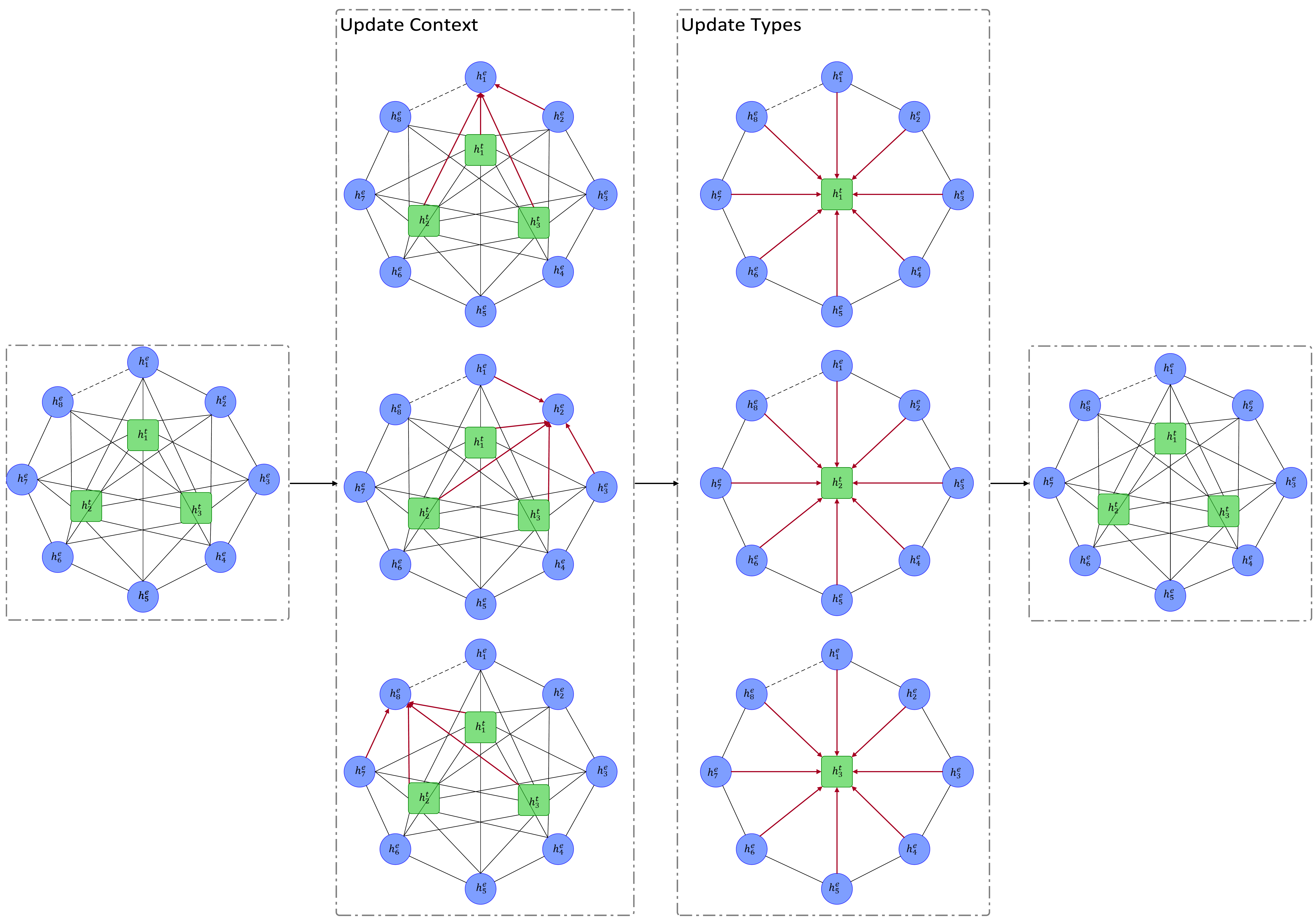}
    \caption{Graph structure and update. The blue circles denote text nodes and the green circles denote type vertices. The red arrow line indicates the direction of the iterative update. Each dashed box corresponds to an update step. The dashed line connecting two nodes indicates that the two nodes are not actually connected.}
    \label{graph}
\end{figure*} 
Figure \ref{graph} illustrates the created heterogeneous graph topology.
Each type node in the graph connects to all text nodes. 
Each text node links to all type nodes and adjacency text nodes.

In terms of type nodes, the graph is bipartite.
As type nodes can see the whole context, this structure allows the converging of global information.
The graph is similar to a convolution network with respect to text nodes.
This structure facilitates text nodes to integrate local information.
It also implicitly introduces relative position, as the text nodes only communicate with text nodes within a fixed-size window.
And global information provided by type nodes helps distinguish between the entity node and the background node.
It can be markedly beneficial for the detection of entity boundaries.

The star-connection topology also achieves a significant reduction in computational complexity.
The common self-attentive mechanism is equivalent to a fully connected homogeneous graph with a theoretical computational complexity of $O(N^2)$.
In contrast, the theoretical computational complexity of our star graph is approximate $O(N)$ due to the limitation on the number of nodes each node passes message with.
Due to its regularity, the star graph is easy to achieve parallelism in implementation.
The parallelism in implementation is essential to convert theoretical complexity advances into practical performance gains.

\subsubsection{Iteration}
The graph nodes are iteratively updated to obtain a semantically richer representation.
One iteration process consists of two parts: the first part update text nodes $H^e$ , and the second part updates the type node $H^y$.
Each part follows the steps below.

\paragraph{\textbf{Transformation}}
For node $j$, linear transformation is applied to project the representation $h_j^l$ of the $l$-th layer $j$ nodes into the semantic feature space of a particular node type $\mu$.
\begin{equation}
    \hat{h}^{(l+1,m)}_{\mu,j}=W^{(l+1,m)}_{\mu,\phi(j)}h_j^l+b^{(l+1,m)}_{\mu,\phi(j)}
\end{equation}
where $W^{(l+1,m)}_{\mu,\phi(j)}\in \mathbb{R}^{n^{(l+1,m)}_{\mu}\times n^{k}_{\phi(j)}}$, $b^{(l+1,m)}_{\mu,\phi(j)} \in \mathbb{R}^{n^{(l+1,m)}_{\phi(j)}}$.
$\hat{h}^{(l+1,m)}_{\mu,j}\in \mathbb{R}^{n^{(l+1,m)}_{\phi(j)}}$ is the representation after mapping previous hidden state from the feature space of type $\phi(j)$ to the feature space of type $\mu$ in the $l+1$-th layer, $m$-th head.

Obviously, the transformation parameters $W^{(l+1,m)}$ and $b^{(l+1,m)}$ has a total of ${|\mathcal{T}|}^2$ in the $l$-th layer, $m$-th head.
To aggregate the neighboring nodes set $\mathcal{N}_i$ of any node $i$, the representation of any node $j$ in the set should be first projected into the semantic feature space of type $\phi(i)$.

\paragraph{\textbf{Aggregation}}
In this paper, we utilize the graph attention mechanism to pass messages between nodes.
In order to take the direction of the edges into account, we design the attention score function as follows:
\begin{equation}
    \begin{aligned}
        s_{i,j}&=\sigma(g_{i,j} + p_{i,j})\\
        g_{i,j}&=a^T(\hat{h}_{\phi(i),\phi(j)}\Vert \hat{h}_{\phi(i),\phi(i)})\\
        p_{i,j}&=\hat{h}^T_{\phi(i),\phi(j)} W_p \hat{h}_{\phi(i),\phi(i)}
    \end{aligned}
    \label{hybrid_attention}
\end{equation}
where $\sigma$ stands for the activate function, $g_{i,j}$ and $p_{i,j}$ denote the different parts of the score. 
We exploit \text{LeakyRelu} \cite{maas2013rectifier} as activate function here as most graph attention networks do.
Note that we have omitted the complex superscript here.
This process of computing attention is actually carried out separately for each attention head at every layer.

Many graph attention networks, such as \cite{velivckovic2017graph}, employ a similar approach to compute $s_{ij}$ as follows:
\begin{equation}
    \begin{aligned}
        \hat{h}^l_i&=W h_i^l\\
        s_{i,j}&=\text{LeakyRelu}(a^T(\hat{h}^l_i\Vert\hat{h}^l_j))
    \end{aligned}
    \label{grapa_attention}
\end{equation}
However, this approach is not appropriate enough for the heterogeneous graph network in our paper. 
Equation \ref{grapa_attention} can be reformulated  as follows:
\begin{equation}
    \begin{aligned}
        a&=a_u\Vert a_d\\
        s_{i,j}&=\text{LeakyRelu}(a^T_u\hat{h}_j+a^T_d\hat{h}_j)
    \end{aligned}
    \label{grapa_attention_change}
\end{equation}
Denote $s_i=[s_{i,j_1},s_{i,j_1},...,s_{i,j_N}]$, where $N=|\mathcal{N}_i|$.
It can be obviously concluded from equation \ref{grapa_attention_change} that $s_i$ and $s_{i^{\prime}}$ should be order-preserving (stay the relative size relationship as the same) if $\mathcal{N}_i = \mathcal{N}_{i^{\prime}}$. 
The reason for this phenomenon is that $a^l_d\hat{h}_j$ in equation \ref{grapa_attention_change} keep the same among the neighborhood sets $\mathcal{N}_i$ and $\mathcal{N}_{i^{\prime}}$.
So for different $i$, $a^l_d\hat{h}_i$ is equivalent to different constant added to the score distribution.
Since $\text{LeakyRelu}$ is a monotonic function, the distribution would keep approximately identical.

The order-preserving nature of $s_i$ and $s_{i^{\prime}}$ implies that in a bipartite-like graph, one side of the nodes shares similar score distributions.
In our graph, this phenomenon means that each type node has a roughly same distribution of attention weights for text nodes, which could be somehow unreasonable since each type node should pay attention to different parts of the context.

From the equation \ref{grapa_attention}, we claim that the concatenate-based attention score function does not directly take into account the connection between the two hidden vectors.
Rather than that, it actually separately score the importance of two hidden vectors and add them up.
Under this observation, we further extend the scoring function into hybrid form as equation \ref{hybrid_attention}.
The hybrid scoring function is a combination of the graph attention and the common QKV attention \cite{2017Attention} while retaining the characteristics of both. 
The hybrid attention mechanism considers both the importance of nodes individually and the closeness of interconnection between nodes.

In addition, equation \ref{hybrid_attention} implicitly represent the direction of edges as the computation of $g_{i,j}$ and $p_{i,j}$ are both asymmetric. 
And when the $i$ and $j$ are the one node, it also naturally represents the self-connected edges as the order of $i$ and $j$ will not affect the final score.  

Softmax normalization is employed on $s_{i,j}$ to yield the attention weights:
\begin{equation}
    \alpha^{(l+1,m)}_{i,j}=\frac{e^{s^{(l+1,m)}_{i,j}}}{\sum\limits_{n\in \mathcal{N}}e^{s^{(l+1,m)}_{i,n}}}
\end{equation}
where $\mathcal{N}$ is the aggregated neighboring set. 
It should be noted that for $\phi(i)=t$, the $\mathcal{N}$ is $\overline{\mathcal{N}_i}$ which is just all text nodes.
While for $\phi(i)=e$, the $\mathcal{N}$ is $\mathcal{N}^k_i$ (without consider the edge between text nodes and type nodes).
It means each text node collects messages from every type node and the text nodes within a window size.
Here $k$ is a hyperparameter determining the size of the receptive field of the text nodes.

With the attention weights, the neighbor nodes are aggregated by weighted sum:
\begin{equation}
    g_i^{(l+1,m)}=\sum_{j\in \mathcal{N}}\alpha_{i,j}^{(l+1,m)}\hat{h}^{(l+1,m)}_{\phi(i),j}
\end{equation}

The final aggregated results will be the concatenation of different heads.
\begin{equation}
g_i^{(l+1)}= \mathop{\parallel}_{m=1}^M g_i^{(l+1,m)}
\end{equation}

\paragraph{\textbf{Update}}
We exploit the gate mechanism in the GRU to update the node state.
\begin{equation}
    \begin{aligned}
        r &= \text{Sigmoid}(W_{xr}x+b_{xr}+W_{hr}h+b_{hr})\\
        z &= \text{Sigmoid}(W_{xz}x+b_{xz}+W_{hz}c+b_{hz})\\
        n &= \text{Tanh}(W_{xn}x+b_{xn}+r\odot(W_{hn}h+b_{hn}))\\
        o &= (1-z)\odot n+z\odot h\\\\
    \end{aligned}
\end{equation}
where $x$ is the current input, $h$ is the hidden state, and $o$ is the final new state output.

\begin{figure}
    \centering   
    \subfigure[Update type nodes.]
    {
    	\includegraphics[width=0.225\textwidth]{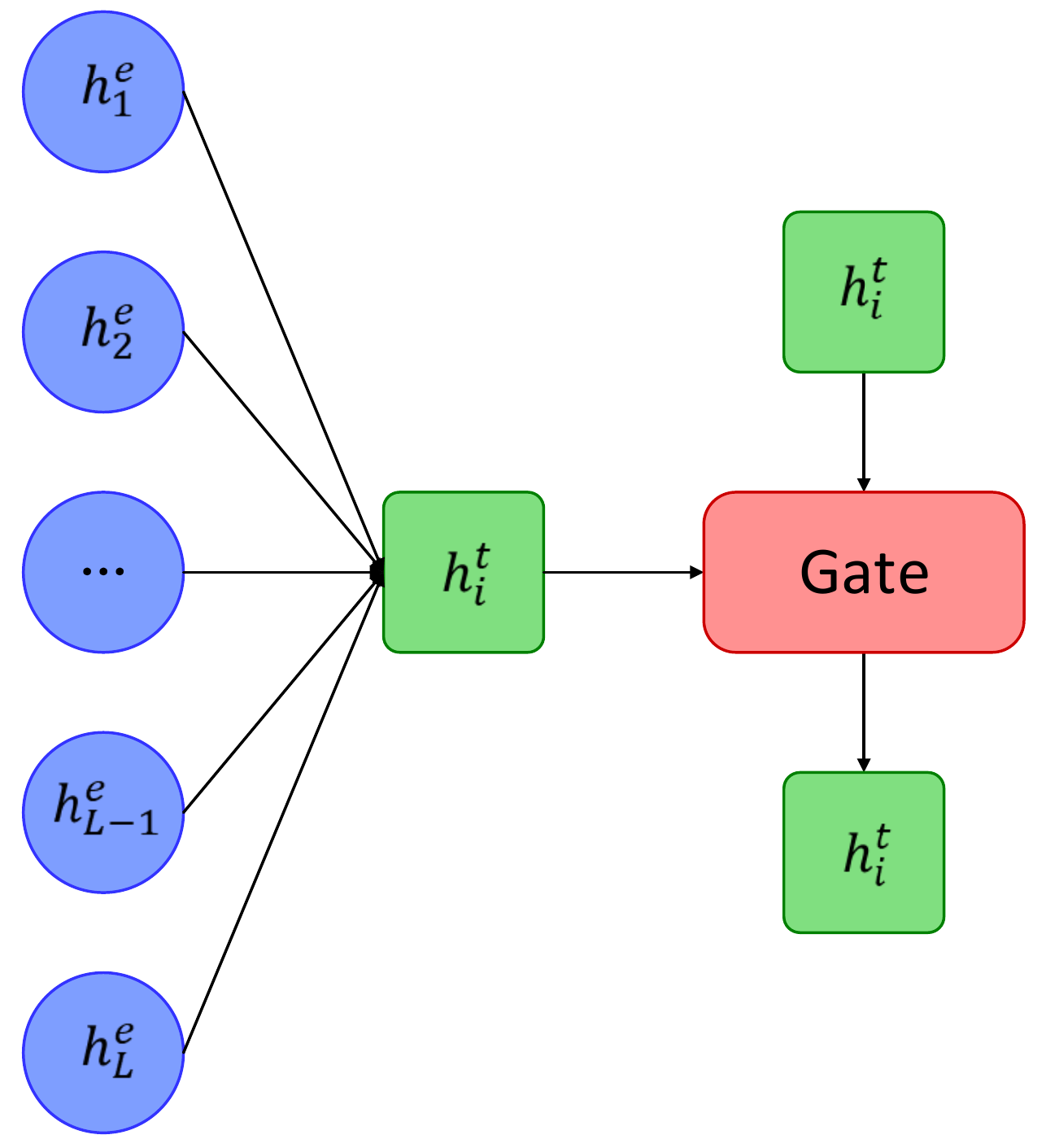} 
    }
    \subfigure[Update text nodes.]
    {
    	\includegraphics[width=0.225\textwidth]{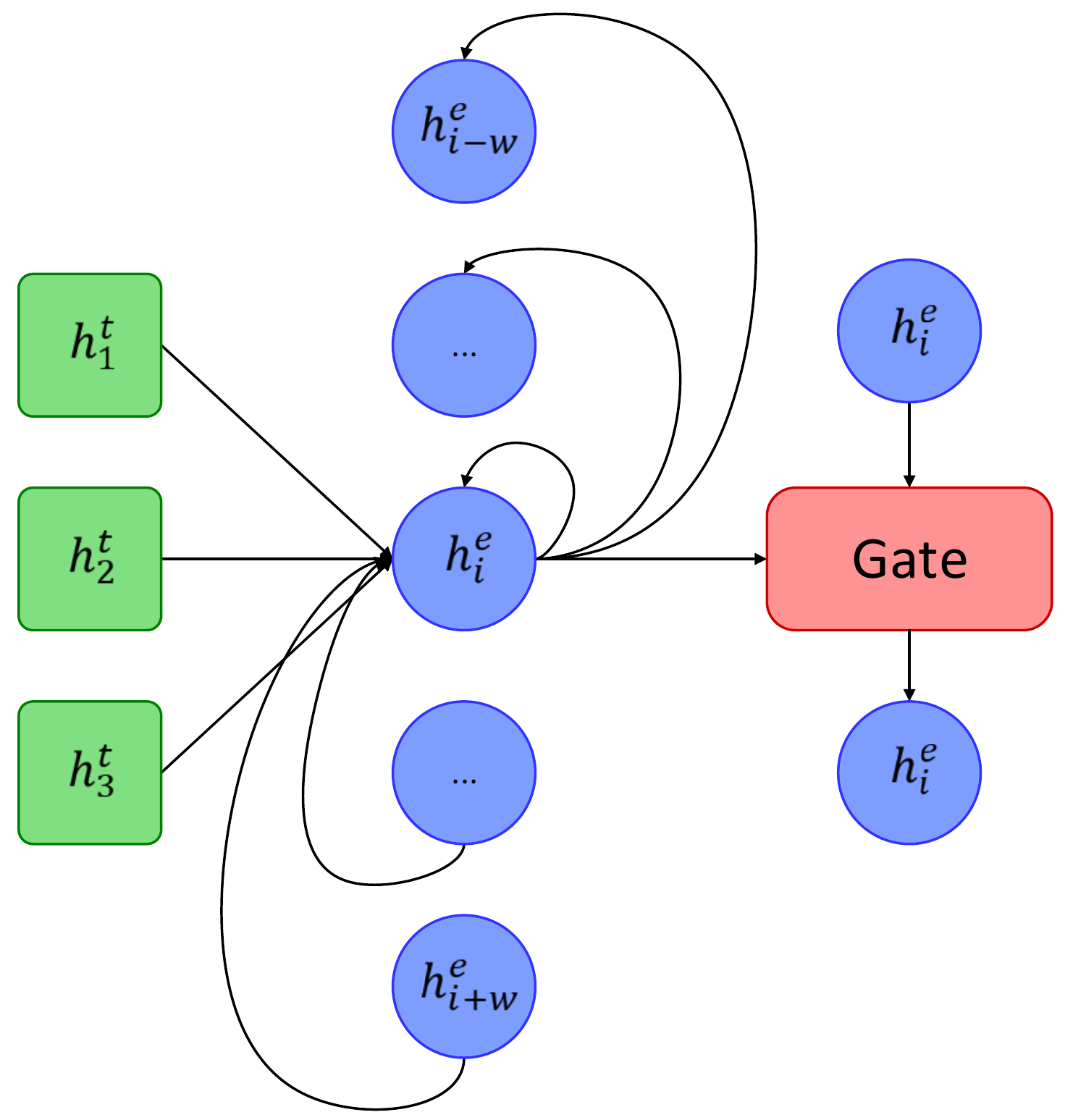}
    }
    \caption{Update of nodes. Type node selectively aggregates information from all text nodes. Text nodes aggregate global and local information from neighboring nodes within a certain window size and all type nodes.}
    \label{update}
\end{figure}

Denote the above mechanism as $\text{Gate}(\cdot)$.
After getting the aggregated result $g_i^{(l+1)}$, the hidden state of nodes is updated as follows:
\begin{equation}
h_i^{(l+1)}= \text{Gate}(h_i^{(l)}, g_i^{(l+1)})
\end{equation}
One iteration consists of two steps of aggregation and update.
The first step is done on the text nodes $H^e$.
The next step is done on the type nodes $H^t$.
The gate mechanism of the two update steps does not share parameters, i.e., it corresponds one-by-one with the node type, as in Figure \ref{update}.

\subsection{Entity Recognition}
Once obtaining the last representation of each node, we perform the named entity recognition based on it.
The following explains how to do the type-supervised sequence labeling with the type nodes and text nodes.

\subsubsection{Fusion}
We combine each type node $h^t$ with the text nodes' representations $H^e$.
The text nodes' representations $H^e$ are first sent to the BiGRU network with the type node hidden states $h^t$ as the initial hidden state:
\begin{equation}
    \begin{aligned}
        H^{\prime}_t &= \text{BiGRU}(H^e, h^t)\\
                     &= [h^{\prime}_{t,1},h^{\prime}_{t,2},...,h^{\prime}_{t,L}]
    \end{aligned}
\end{equation}

The outputted hidden states $H^{\prime}_t$ is the fused text representation:
Subsequently, each fused text hidden states is further integrated with the node representations as follows:
\begin{equation}
    \begin{aligned}
        h_{t,i} &= \text{Maxout}(h^{\prime}_{t,i} + h^e_i, h^t)\\     
        H_t &= [h_{t,1}, h_{t,2},..., h_{t,L}]
    \end{aligned}
\end{equation}
where \text{Maxout} is the activate function in \cite{goodfellow2013maxout}.

All the entities detected from the text representation $H_t$ are of type $t$ and denoted as  $\mathcal{X}_{t}$.
So the cluster of all entities in the text is the union of every $\mathcal{X}_{t}$.
\begin{equation}
    \mathcal{X}=\mathop{\cup}\limits\limits_{t\in T} \mathcal{X}_{t}
\end{equation}
where $T$ represents the set of all predefined entity categories.

\subsubsection{Multi-layer labeling}

\label{tagging}
BIOES(e.g., B-begin, I-inside, O-outside, E-end, S-single) labeling scheme is a common labeling method in traditional NER.
In this paper, we use the special conditional random field \cite{lafferty2001conditional} to implement BIOES sequence labeling.

For a fused text representation $H_t$, we compute every token's scores for each tag by the linear transformation:
\begin{equation}
    P=W_t H_t+b_t
\end{equation}
where $P\in \mathbb{R}^{L\times G}$ is the score matrix, $G$ is the number of tags.
So the $P_{i,j}$ is the score of the $i$-th token labeled as $j$-th tag.

For the correct sequence of labels $Y=(y_1,y_2,... ,y_L)$, define its score as:
\begin{equation}
    s(P,Y)=\sum_{i=0}^LA_{y_i,y_{i+1}}+\sum_{i=1}^LP_{i,y_i}
\end{equation}
where $A$ is the transition matrix.
$A_{i,j}$ denotes the score of the tag $i$ transmit to the tag $j$. $y_0$ and $y_n$ are the start and end tags of the sequence.
The probability of the correctly labeled sequence is calculated by doing a $\text{Softmax}$ operation on all possible sequences' scores:
\begin{equation}
    p(Y|P)=\frac{e^{s(P,Y)}}{\sum\limits_{\tilde{Y}\in Y_P}e^{s(P,\tilde{Y})}}
\end{equation}
In the training phase, we minimized the negative log probability of the correct label.
\begin{equation}
    -log(p(Y|P))=\mathop{logsum}\limits_{\tilde{Y}\in Y_P}s(P,\tilde{Y})-s(P,Y)
\end{equation}
where $Y_P$ indicate all possible sequences of labels.

In the decoding stage, the sequence with the largest probability is outputted:
\begin{equation}
    Y^*=\mathop{argmax}\limits_{\tilde{Y}\in Y_P}s(P,\tilde{Y})
\end{equation}

The unique feature of utilizing \text{CRF} in our work is that we set all unreasonable label transmission scores to be negative infinity.
The reasons for doing so are listed as follow:
\begin{itemize}
    \item The parameters of the model backbone are considerably larger compared to the CRF. It will result in CRF not being able to learn a reasonable transition matrix $A$. This is because the backbone of the model will reach the fitting state before the CRF is properly learned.
    \item Setting the impossible tag transition score to be $-\infty$ can avoid the emergence of invalid and ambiguous annotations which confuse the decoding algorithm. E.g., the transmission score from \emph{Begin} to Out is set to $-\infty$, which causes all \emph{Begin} tags to either end with a \emph{Single} or an \emph{End} instead of being unclosed(end up with \emph{Inside}).
\end{itemize}

\begin{figure}
	\centering
	\includegraphics[width=0.47\textwidth]{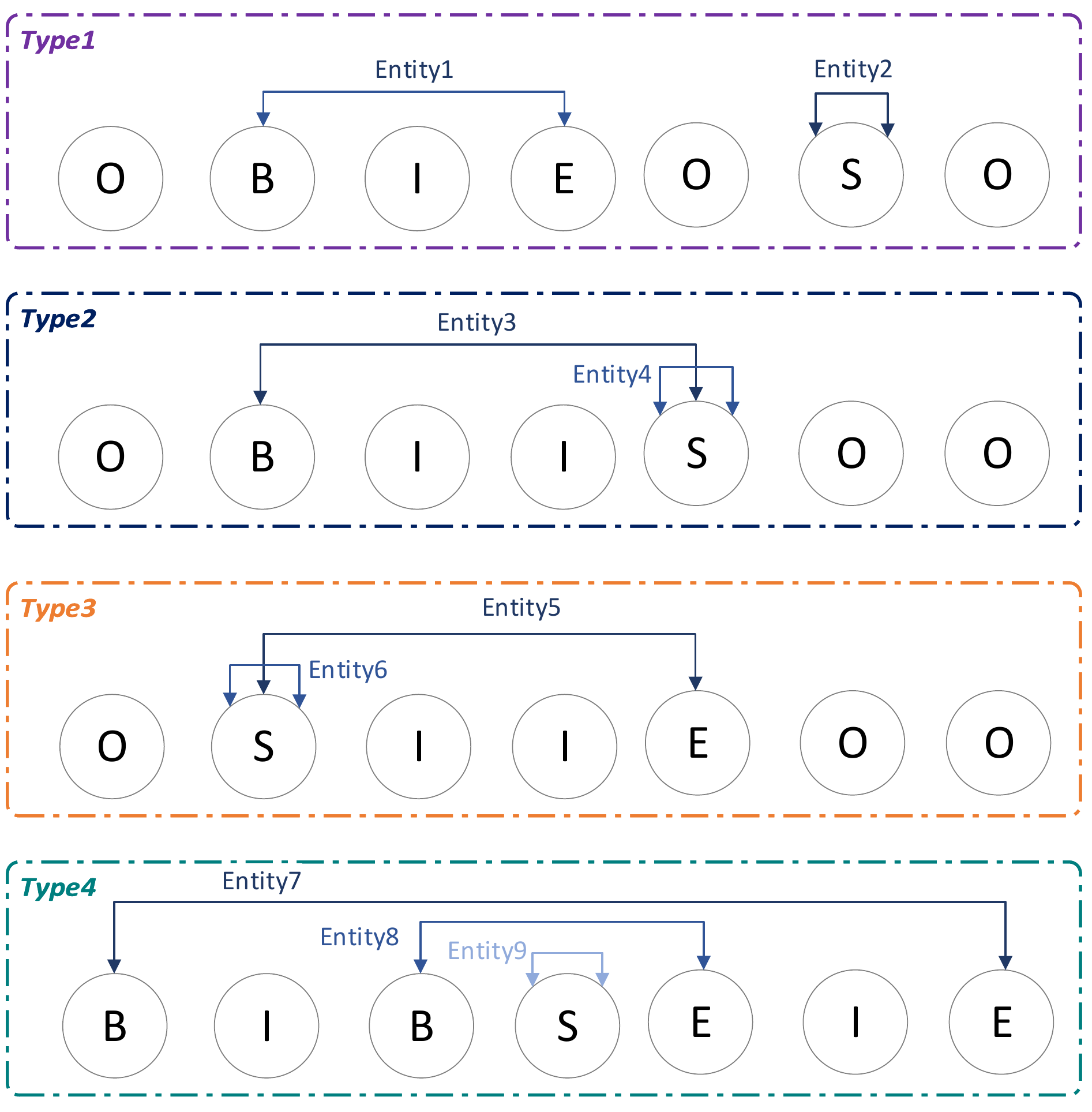}
	\caption{Example of nested BIOES labels. The dashed boxes of different colors indicate the BIOES annotations under different entity categories.}
	\label{nest_bioes_example}
\end{figure}

To identify \textbf{NST}, we extend the normal BIOES labeling scheme and its decoding algorithm. 
A typical nested BIOES annotation in our work is shown in Figure \ref{nest_bioes_example}.
As can be seen, there are several kinds of nested entities in the example, such as Entity1 and Entity3 are \textbf{NDT}, Entity5 and Entity6 are \textbf{NST} and  Entity3 and Entity5 are \textbf{ME}. 

For the annotation of nested entity sets, the algorithm \ref{nested_tagging} is performed. 
It labels entities in the order of \emph{Inside} tag first, \emph{End}, \emph{Begin} second, and \emph{Single} last.
The order allows for maximum preservation of the hierarchical relationships of nested entities when only using BIOES annotations.

\begin{algorithm}
	\caption{Nested BIOES tagging}
	\label{nested_tagging}
	\begin{algorithmic}[1]
	    \STATE \textbf{Input}: Set of entities $\mathcal{X}_t$ of type $t$
		\STATE $T[0:L] \gets O$
		\FOR{$(s,e,t)$ \textbf{in} $\mathcal{X}_t$}
		    \STATE $T[s+1:e-1] \gets \text{I}$
		\ENDFOR
        \FOR{$(s,e,t)$ \textbf{in} $\{(s,e,t)|(s,e,t)\in \mathcal{X}_t \wedge s\ne e\}$}
            \STATE $T[s] \gets \text{B};T[e] \gets \text{E}$
		\ENDFOR
		\FOR{$(s,e,t)$ \textbf{in} $\{(s,e,t)|(s,e,t)\in \mathcal{X}_t \wedge s=e\}$}
	        \STATE $T[s] \gets \text{S}$
        \ENDFOR
	\end{algorithmic}  
\end{algorithm}

And for decoding nested BIOES annotations correctly, the decoding algorithm \ref{nested_detagging} is given based on the principle of decoding entities from the inside out.
The algorithm first converts all the tags into a queue of start and end indexes.
The algorithm enumerates the entity index pairs in the queues.
When the entity pair is already at the innermost level or not nested by any entity, it is considered to constitute an entity and then popped. 
When an entity is popped, the algorithm will decide whether its start or end index may from another entity, and change the tags accordingly. 

\begin{algorithm}
	\caption{Nested BIOES Detagging}
	\label{nested_detagging}
	\begin{algorithmic}[1]
	    \STATE \textbf{Input}: Sequence of BIOES tags $T$ of type $t$
		\STATE $S \gets \{ i|(T[i]=\text{B})\vee( T[i]=\text{S})\}$
		\STATE $E \gets \{i|(T[i]=\text{E})\vee( T[i]=\text{S})\}$
		\STATE $Ent \gets \{\}$
		\WHILE{$|S\times E|>0 $}
    		\FOR{$(s,e)$ \textbf{in} $S\times E$}
        	    \IF{$(s,e)\text{ form an innermost entity}$} 
        	        \STATE $Ent \gets Ent+{(s,e,t)}$
                    \IF{no more $(s,e^{\prime})$ form an entity} 
                        \STATE S.pop(s); 
                    \ELSE
                        \STATE $T[s] \gets \text{B}$
                    \ENDIF 
                    \IF{no more $(s^{\prime},e)$ form an entity}
                        \STATE E.pop(e);
                    \ELSE
                        \STATE $T[e] \gets \text{E}$
                    \ENDIF 
                    \IF{$(s,e)$ totally nested}
                        \STATE $T[s] \gets \text{I}; [e] \gets \text{I}$
                    \ENDIF 
                    \STATE \textbf{GOTO} 5
                \ENDIF 
    		\ENDFOR
		\ENDWHILE
	\end{algorithmic}  
\end{algorithm}
\section{Experiments}
In this section, we detail the datasets, baselines, and settings used in the experiments.
We conducted several experiments to study the properties of our model.
In addition, we investigate the cases in which our method achieved satisfactory results and the scenarios in which our model failed.

\subsection{Dataset and Evaluation}
This section will describe the three public datasets used for our experiments. 
We evaluate our model on nested datasets GENIA \cite{kim2003genia}, flat dataset CoNLL-2003\ cite{sang2003introduction} and Chinese dataset Weibo NER \cite{peng2015named}. 
The statistical information of the data is shown in Table \ref{statistic}.

GENIA is a nested dataset in the biological field containing five entity types, including DNA, RNA, protein, cell lineage, and cell type. 
We keep the same experiment setting with \cite{yu-etal-2020-named}.
We did not conduct experiments on other authoritative English nested datasets, mainly because we failed to obtain the copyright of these datasets.

The CoNLL-2003 dataset is a large dataset widely used by NER researchers.
Its data source is the Reuters RCV1 corpus, whose main content is news reports. 
Its named features include locations, organizations, people, and information.
We follow the practice of \cite{yan2021unified} to merge the training and validation sets.
Our experiments on this dataset demonstrate our model's generalizability on the flat dataset.

Weibo NER is a Chinese dataset sampled from the Weibo social platform with four types of entities, including personal, organizational, location, and geo-political. 
Each type is divided into name and nominal mentions.
Our experiments on this dataset share the same settings with \cite{XiaonanLi2020FLATCN}.
We utilize this Chinese dataset to verify the cross-linguistic ability of the model.

\begin{table*}
\centering
\caption{Statistical information on datasets.}
\label{statistic}
\begin{tabular}{lccccccccc}
\hline
\multirow{2}{*}{Info.} & \multicolumn{3}{c}{GENIA} & \multicolumn{3}{c}{CoNLL-2003} & \multicolumn{3}{c}{WeiboNER} \\ \cline{2-10} 
                    & Train & Test & Dev  & Train & Test & Dev  & Train & Test & Dev  \\ \hline
Sentences           & 15203 & 1669 & 1854 & 14041 & 3250 & 3453 & 1350  & 270  & 270  \\
Avg sentence length & 25.4  & 24.6 & 26.0 & 14.5  & 15.8 & 13.4 & 33.6  & 33.2 & 33.8 \\
Entities            & 46142 & 4367 & 5506 & 23499 & 5942 & 5648 & 1834  & 371  & 405  \\
Avg entities length & 1.94  & 2.14 & 2.08 & 1.43  & 1.43 & 1.42 & 1.24  & 1.19 & 1.20 \\ \hline
\end{tabular}
\end{table*}

\subsection{Training Details}
In most experiments, we and pre-trained BERT model implemented by framework Transformers \cite{ThomasWolf2019HuggingFacesTS}.
For the GENIA dataset, we replaced BERT with BioBERT  \cite{10.1093/bioinformatics/btz682} in order to obtain a better contextual representation in biological domain text.
For English dataset, we utilize GloVE \cite{pennington-etal-2014-glove} as pre-trained word vectors.
For Chinese dataset, we exploit word vectors created by \cite{P18-2023}.
In all experiments for comparison, we set $d^C = 100$, $d^K=1024$ and $d^P=25$.
For all experiments, the models are trained with the \text{AdamW} optimizer with learning rate equals to $2e-5$ and max gradient normalization equals to $1e0$.
Learning rate are altered during the training by the linear warm-up-decay learning rate schedule.

\subsection{Comparison}
For GENIA dataset, we compare our model with several powerful state-of-the-art models, including \textbf{LocateAndLabel}\cite{shen2021locate}, \textbf{SequenceToSet}\cite{ZeqiTan2021ASN},  \textbf{BARTNER} \cite{HangYan2021AUG}, \textbf{LogSumExpDecoder}\cite{YiranWang2021NestedNE}.
The reported results of the above baselines are directly transferred from the original published literature.

For the CoNLL-2003 dataset, we make the comparison with \textbf{LocateAndLabel}\cite{shen2021locate}, \textbf{BARTNER}\cite{HangYan2021AUG}, \textbf{NER-DP}\cite{yu2020named}, \textbf{LUKE}\cite{yamada-etal-2020-luke}, \textbf{MRC}\cite{li-etal-2020-unified} .
We copy the results from the original published literature.
Note that the outcomes of LUKE\cite{yamada-etal-2020-luke} and NER-DP\cite{yu2020named}, MRC\cite{li-etal-2020-unified} are obtained from \cite{HangYan2021AUG}.

For Weibo NER dataset, we compete our model with several strong baselines, namely \textbf{LocateAndLabel}\cite{shen2021locate}, \textbf{SLK-NER}\cite{DouHu2020SLKNERES}, \textbf{FLAT}\cite{XiaonanLi2020FLATCN}, \textbf{SANER}\cite{DouHu2020SLKNERES}, \textbf{AESINER}\cite{YuyangNie2020ImprovingNE}.
We report results demonstrated in the original published literature.

Table \ref{result} illustrate the results of different methods for named entity recognition on GENIA, Conll2003, and Weibo NER datasets. 
The outcomes show that our method has advanced performance. 
Our model reaches a 0.17\% improvement in F1 score over the best method in the GENIA dataset. 
Since the GENIA dataset contains a high portion of nested entities, we believe the excellent performance on this dataset proves the ability of our model to extract the nested entity.
Additionally, the significant progress in precision on the three evaluated datasets reflects the superiority of the proposed type-supervised sequence labeling method.

\begin{table}
\caption{The overall performance.}
\centering
\label{result}
\begin{tabular}{llll}
\toprule
\multirow{2}{*}{Model}                       & \multicolumn{3}{c}{GENIA}                                 \\ \cline{2-4} 
                                             & Prec.            & Rec.              & F1                 \\ \hline
Pyramid\cite{wang-etal-2020-pyramid}         & 79.45            & 78.94             & 79.19              \\
NER-DP\cite{yu2020named}                     & 81.80            & 79.30             & 80.50              \\
LocateAndLabel\cite{shen2021locate}          & 80.19            & \textbf{80.89}    & 80.54              \\
SequenceToSet\cite{ZeqiTan2021ASN}           & 82.31            & 78.66             & 80.44              \\
BARTNER\cite{HangYan2021AUG}                 & 78.89            & 79.60             & 79.23              \\
LogSumExpDecoder\cite{YiranWang2021NestedNE} & 79.20            & 78.67             & 78.93              \\ \hline
Our Model                                    & \textbf{83.02}   & 78.53             & \textbf{80.71}     \\ \bottomrule
\toprule
\multirow{2}{*}{Model}                       & \multicolumn{3}{c}{CoNLL-2003}                          \\ \cline{2-4} 
                                             & Prec.               & Rec.           & F1               \\ \hline
LUKE\cite{yamada2020luke}                    & -                   & -              & 92.87            \\
MRC\cite{li-etal-2020-unified}               & 92.47               & 93.27          & 92.87            \\
NER-DP\cite{yu2020named}                     & 92.85               & 92.15          & 92.50            \\
LocateAndLabel\cite{shen2021locate}          & 92.13               & 93.79          & 92.94            \\
BARTNER\cite{HangYan2021AUG}                 & 92.61               & \textbf{93.87} & 93.24            \\ \hline
Our Model                                    & \textbf{93.26}      & 93.23          & \textbf{93.25}   \\ \bottomrule
\toprule
\multirow{2}{*}{Model}                       & \multicolumn{3}{c}{Weibo NER}                            \\ \cline{2-4} 
                                             & Prec.              & Rec.            & F1                \\ \hline
FLAT\cite{XiaonanLi2020FLATCN}               & -                  & -               & 68.55             \\
SLK-NER\cite{DouHu2020SLKNERES}              & 61.80              & 66.30           & 64.00             \\ 
SANER\cite{DouHu2020SLKNERES}                & -                  & -               & 69.80             \\ 
AESINER\cite{YuyangNie2020ImprovingNE}       & -                  & -               & 69.78             \\ 
LocateAndLabel\cite{shen2021locate}          & 70.11              & \textbf{68.12}  & 69.16             \\ \hline
Our Model                                    & \textbf{74.44}     & 66.17           & \textbf{70.06}    \\ \bottomrule
\end{tabular}
\end{table}

\subsection{Detailed Result}
We demonstrate the result on different kinds of nested entities on Table \ref{nested_result}.
It can be concluded that our proposed method can achieve consistent accuracy on general entities and nested entities.
And we believe that the relatively low recall in nested entities is caused by the relatively low proportion of nesting in the total entities.

We also analyzed the performance of entities of different lengths as shown in Table \ref{entity_length}. 
Our model achieves the best results when the entity length is $1$ or $2$ on all datasets, which happens to be the window size used in our experiments.
This phenomenon indirectly shows the contribution of the star topology of the graph.

\begin{table}
\centering
\caption{Result on different kinds of nested entity.}
\label{nested_result}
\begin{tabular}{lcccc}
\hline
\multicolumn{1}{c}{\multirow{2}{*}{Category}} & \multicolumn{4}{c}{GENIA}                        \\ \cline{2-5} 
\multicolumn{1}{c}{}                          & Prec. & Rec.  & F1    & \multicolumn{1}{l}{Num.} \\ \hline
NST                                           & 82.26 & 61.60 & 70.44 & 625                      \\
NDT                                           & 84.09 & 67.27 & 74.74 & 605                      \\
Flat                                          & 83.74 & 82.54 & 83.13 & 4307                     \\ \hline
All                                           & 82.85 & 78.47 & 80.60 & 5506                     \\ \hline
\end{tabular}
\end{table}

\begin{table*}
\centering
\caption{Result on entities of different lengths.}
\label{entity_length}
\begin{tabular}{lccccccccc}
\hline
\multicolumn{1}{c}{\multirow{2}{*}{Length}} &
  \multicolumn{3}{c}{GENIA} &
  \multicolumn{3}{c}{CoNLL-2003} &
  \multicolumn{3}{c}{Weibo NER} \\ \cline{2-10} 
\multicolumn{1}{c}{} & Prec. & Rec.  & F1    & Prec. & Rec.           & F1    & Prec. & Rec.  & F1    \\ \hline
$l=1$ &
  \textbf{87.24} &
  77.18 &
  \textbf{81.90} &
  94.12 &
  92.60 &
  93.35 &
  \textbf{84.05} &
  \textbf{70.94} &
  \textbf{76.94} \\
$l=2$ &
  85.24 &
  \textbf{80.57} &
  81.56 &
  \textbf{94.74} &
  95.02 &
  \textbf{94.88} &
  50.70 &
  47.36 &
  48.97 \\
$l=3$                & 78.06 & 77.64 & 77.85 & 87.93 & 89.47          & 88.69 & -     & -     & -     \\
$l=4$                & 77.94 & 78.14 & 78.04 & 94.59 & 94.59          & 94.59 & -     & -     & -     \\
$l\ge 5$             & 73.93 & 78.99 & 76.37 & 88.88 & \textbf{96.00} & 92.30 & -     & -     & -     \\ \hline
All                  & 83.02 & 78.53 & 80.71 & 93.26 & 93.23          & 93.25 & 74.44 & 66.17 & 70.06 \\ \hline
\end{tabular}
\end{table*}

\subsection{Analysis and Discussion}

\subsubsection{Ablation Study}
We examined the contributions of different modules in the model at the GENIA dataset.
First, we replaced the gating mechanism with the single layer perception machine with $\text{Tanh}$ activation function. 
We also investigated the impact of word node update, type node update.
Finally, we use the traditional graph attention instead of hybrid attention to evaluate our proposed mechanism.
Table \ref{ablation_study} shows the outcomes. 

The performance degradation after replacing the gating mechanism proves its usefulness for learning better node representations.
After removing the text node update, the F1 value drops to $79.73$.
It indicates the importance of fusing type nodes and text nodes representations and aggregating local information.
After removing the type node update, the F1 value declined to $79.85$.
It suggests the significance of type node supervision and utilizing global messages.
The deterioration of performance after replacing hybrid attention confirms the effectiveness of our proposed attention mechanism.

\begin{table}
\centering
\caption{Ablation study. '-' means delete the module or replace it with another module.}
\label{ablation_study}
\begin{tabular}{lccc}
\hline
\multicolumn{1}{c}{\multirow{2}{*}{Model}} & \multicolumn{3}{c}{GENIA} \\ \cline{2-4} 
\multicolumn{1}{c}{}                       & Prec.   & Rec.   & F1     \\ \hline
Origin                                     & 82.25   & 78.18  & 80.16  \\
-Gate mechanism                            & 81.90   & 77.91  & 79.85  \\
-Text node update                          & 81.43   & 78.33  & 79.85  \\
-Type node update                          & 81.51   & 78.02  & 79.73  \\
-Hybrid Attention                          & 81.45   & 78.33  & 79.86  \\ \hline
\end{tabular}
\end{table}

\subsubsection{Number of Graph Layers} 
\begin{figure}
	\centering
	\includegraphics[width=0.48\textwidth]{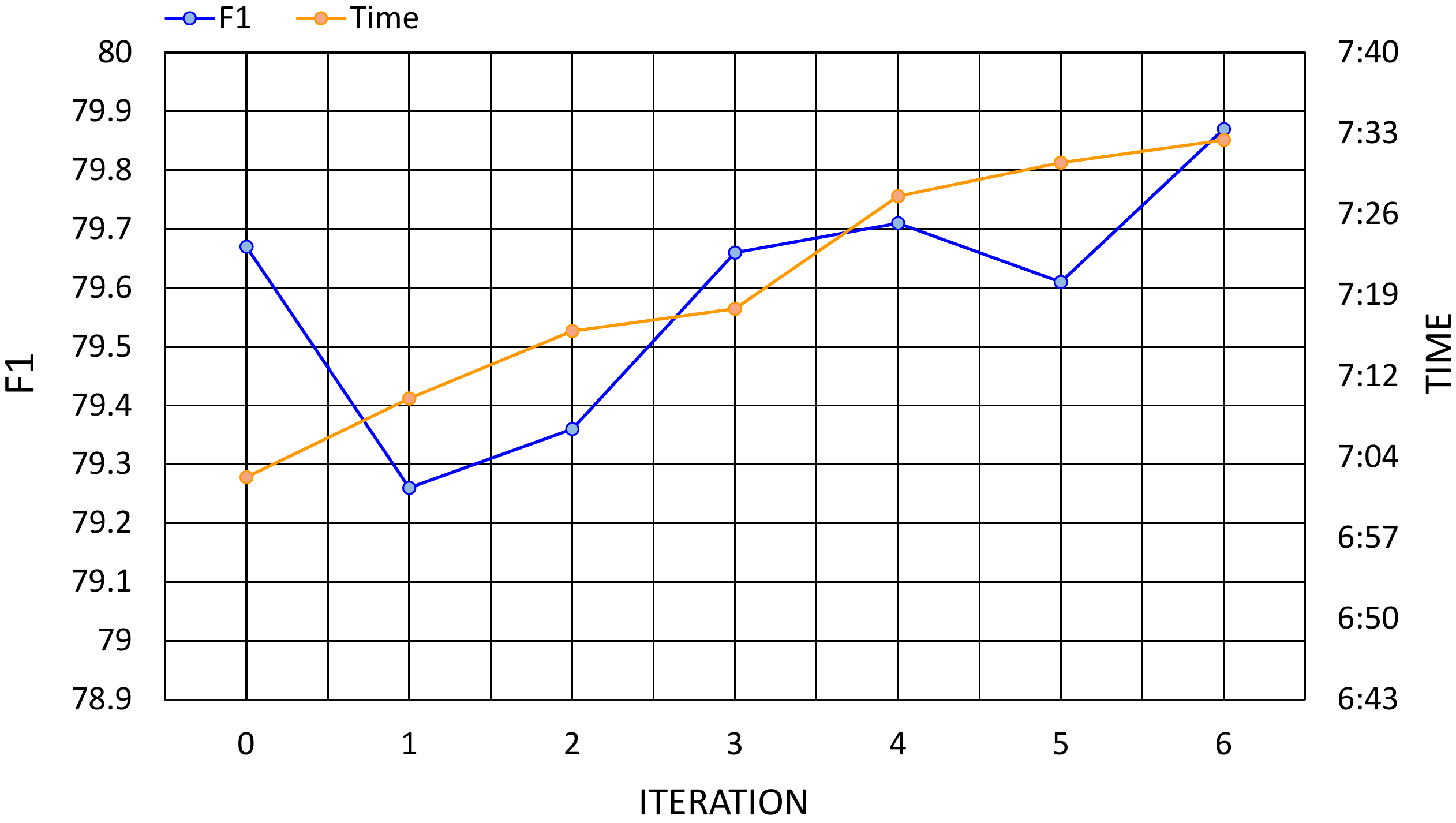}
	\caption{Result on models of different numbers layers.}
	\label{layers}
\end{figure}

To investigate the influence of the layer of our graph neural network, we experimented with models of different layers on the GENIA dataset.
All the models were trained with $5$ epochs and the window size equals $1$.
Table \ref{layers} shows the results for the different layers and corresponding time costs.
We can see our model reaches a comparable result when $l\ge 3$.
To balance the time cost and model performance, we set $l=3$ for other experiments.

\subsubsection{Analysis of Window Size} 
\begin{figure}
	\centering
	\includegraphics[width=0.48\textwidth]{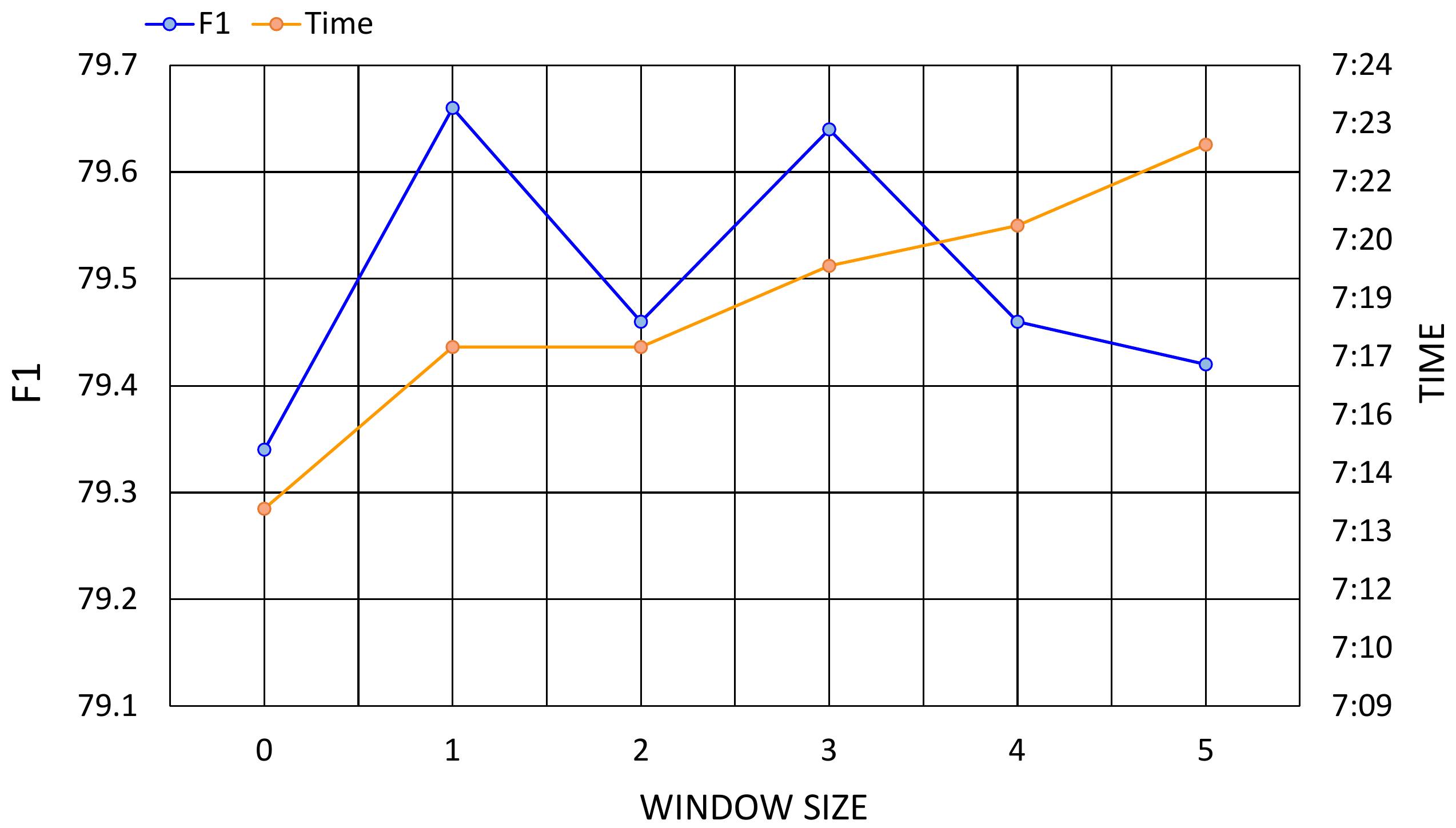}
	\caption{Result of different window sizes of heterogeneous graph network layers. The experiments are performed on the GENIA dataset. Time is the average time per epoch.}
	\label{window_size}
\end{figure}

We also explore the effect of window size in the graph topology, i.e., the size of text nodes' aggregation sets. 
Figure \ref{window_size} reports the relationship between window size and model performance and time cost. 
As the window size increases, the time cost gradually increases. 
However, the model reaches the highest performance when $w=1$.
We believe that this is because the average length of the entity in the GENIA dataset is close to $2$.
In this condition, $w=1$ means the perceptual field of the text nodes at the head and tail positions of most entities exactly cover the whole entity, which will benefit the determination of the entity boundary.
This result also reveals the effectiveness of our proposed star topology from the side.

\subsubsection{Visualization} 
\begin{figure*}[!h]
    \centering   
    \subfigure[Text node present entity.]
    {
    	\includegraphics[width=0.3\textwidth]{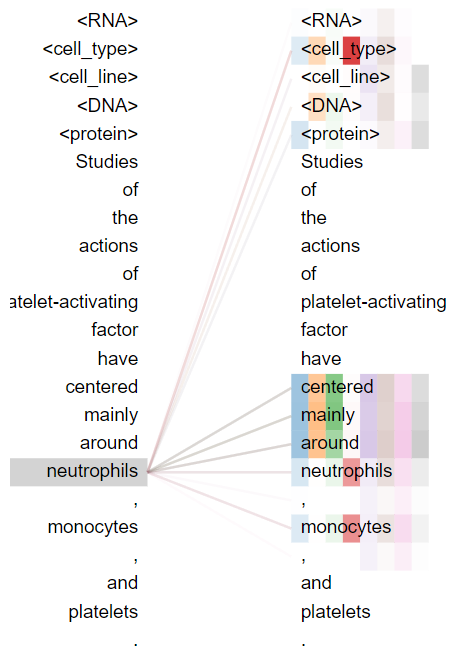} 
    	\label{text_entity_1}
    }
    \subfigure[Text node presenting entity.]
    {
    	\includegraphics[width=0.3\textwidth]{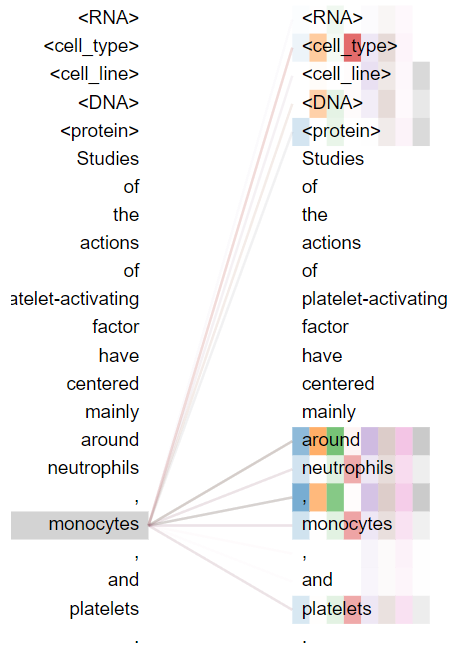}
    	\label{text_entity_2}
    }
    \subfigure[Text node as conjunction.]
    {
    	\includegraphics[width=0.3\textwidth]{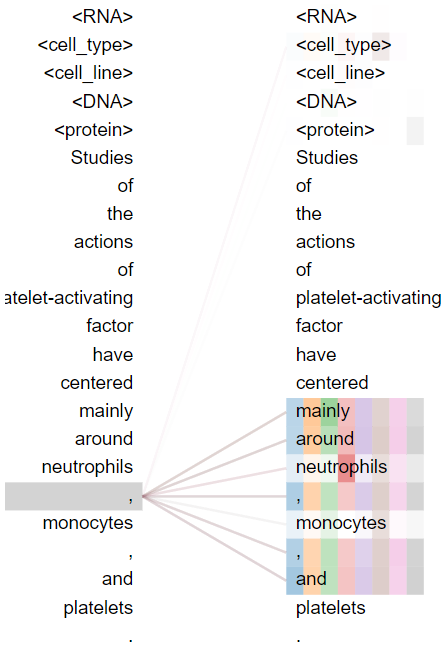}
    	\label{text_conjunction}
    }
    \subfigure[Type nodes presenting entities.]
    {
    	\includegraphics[width=0.3\textwidth]{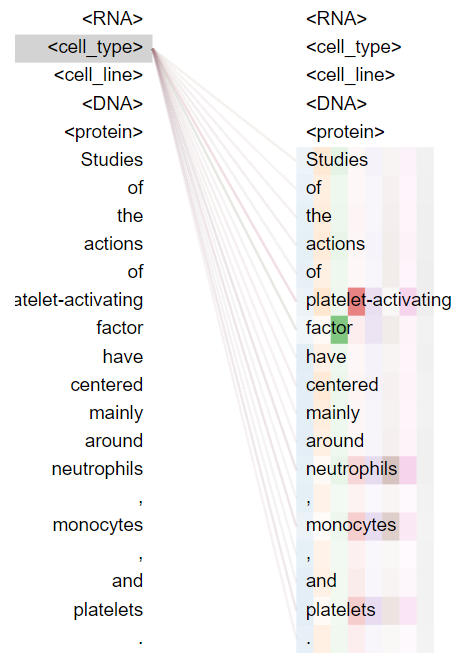} 
    	\label{entity_1}
    }
    \subfigure[Type nodes presenting entities.]
    {
    	\includegraphics[width=0.3\textwidth]{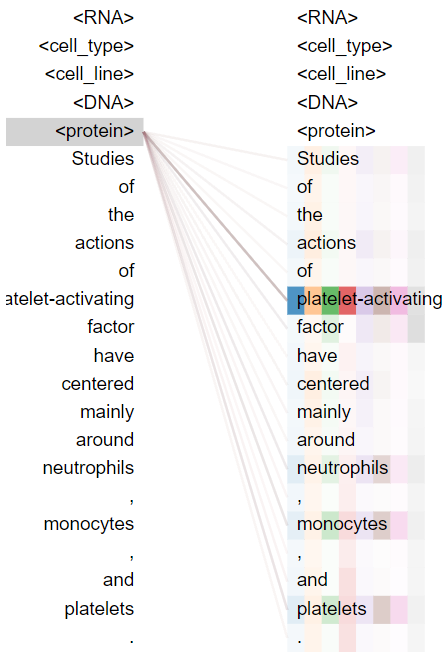}
    	\label{entity_2}
    }
    \subfigure[Type nodes present as global information.]
    {
    	\includegraphics[width=0.3\textwidth]{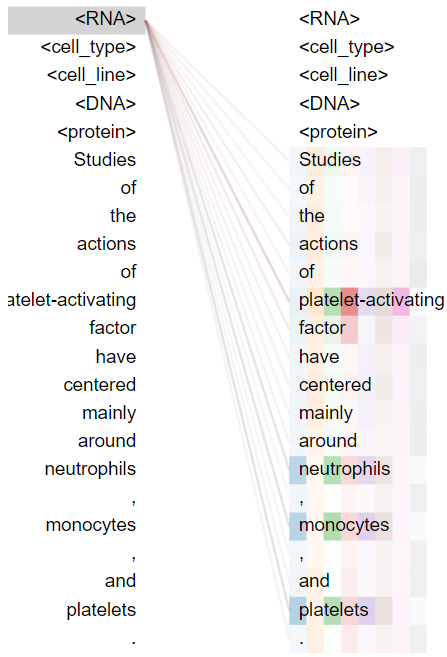}
    	\label{global_type}
    }
    \caption{Attention-head views. The colors in the figure present different heads, and the different transparencies indicate the levels of attention. Words enclosed in sharp brackets denote the category of the entity. The figure is drawn by \cite{vig2019multiscale}.}
    \label{visualiazation}
\end{figure*}

In this section, we visualize the scores of each attention head and show the interpretability and reasonability of our graph attention model.
We use the model trained on the GENIA dataset for visual analysis.
Specifically, we visualize the properly trained heterogeneous graph module's attention scores between the nodes in Figure \ref{visualiazation}.
The top three figures are attention heads' view of the text nodes, and the bottom three figures are types nodes' views.

Analyzing Figure \ref{text_entity_1} and Figure \ref{text_entity_2}, where the text nodes represent the cell-type entities, we can see that the attention heads focus mainly on words with real meaning such as "centered" and ignores the meaningless comma.
Meanwhile, there can be higher attention scores between entities than background words.
It's noteworthy that the comma node \ref{text_conjunction} in between the two entities show similar attention to all background words while having lower attention to entity words.
It's a very different property from entity nodes.
Since the comma node needs to explicitly be a background term to avoid confusion with adjacent entity nodes, the phenomenon of nodes spatially adjacent behaving quite differently is actually understandable.

In Figure \ref{entity_1}, the type node \textit{<cell\_type>} shows higher attention to its three corresponding entities "neutrophils", "monocytes" and "platelets" compared to the other background words. 
At the same time, it also exerts relatively more attention to another entity in the sentence "platelets-activating factor".
And the type node \textit{<protein>} shows a significant focus on its only corresponding entity "platelets-activating factor".
Obviously, the type nodes focus more on the text nodes of the corresponding entities. 
And as for the Figure \ref{global_type}, where the type node \textit{<RNA>} plays the role as a global node, there is a relatively high focus on entity words while a relatively uniform integration of all background words.
Clearly, type nodes' aggregation is selective and effective for extracting information in sentences.

\subsubsection{Complexity}
In our sequence labeling scheme, only one time of annotation is needed for each category of entities.
Thus the time complexity of the decoding part is $O(cN)$, which is an order of magnitude reduction compared with some region-based methods like \cite{shen2021locate} whose theoretical computational complexity is $O(cN^2)$.

As for the attention mechanism, the traditional attention mechanism is approximately equivalent to the fully connected homogeneous graph, whose theoretical computational complexity is $O(N^2)$. 
In this paper, the employment of the star topology reduces the computational complexity to $O(4N+2cN+w^2+2w)$. 
And in the implementation, the vector product after concatenation can be decomposed into the product first and summation later, thus gaining further dismission in space occupation and speedup in computation speed.

\subsubsection{Error Analysis}
\begin{figure}[H]
	\centering
	\includegraphics[width=0.47\textwidth]{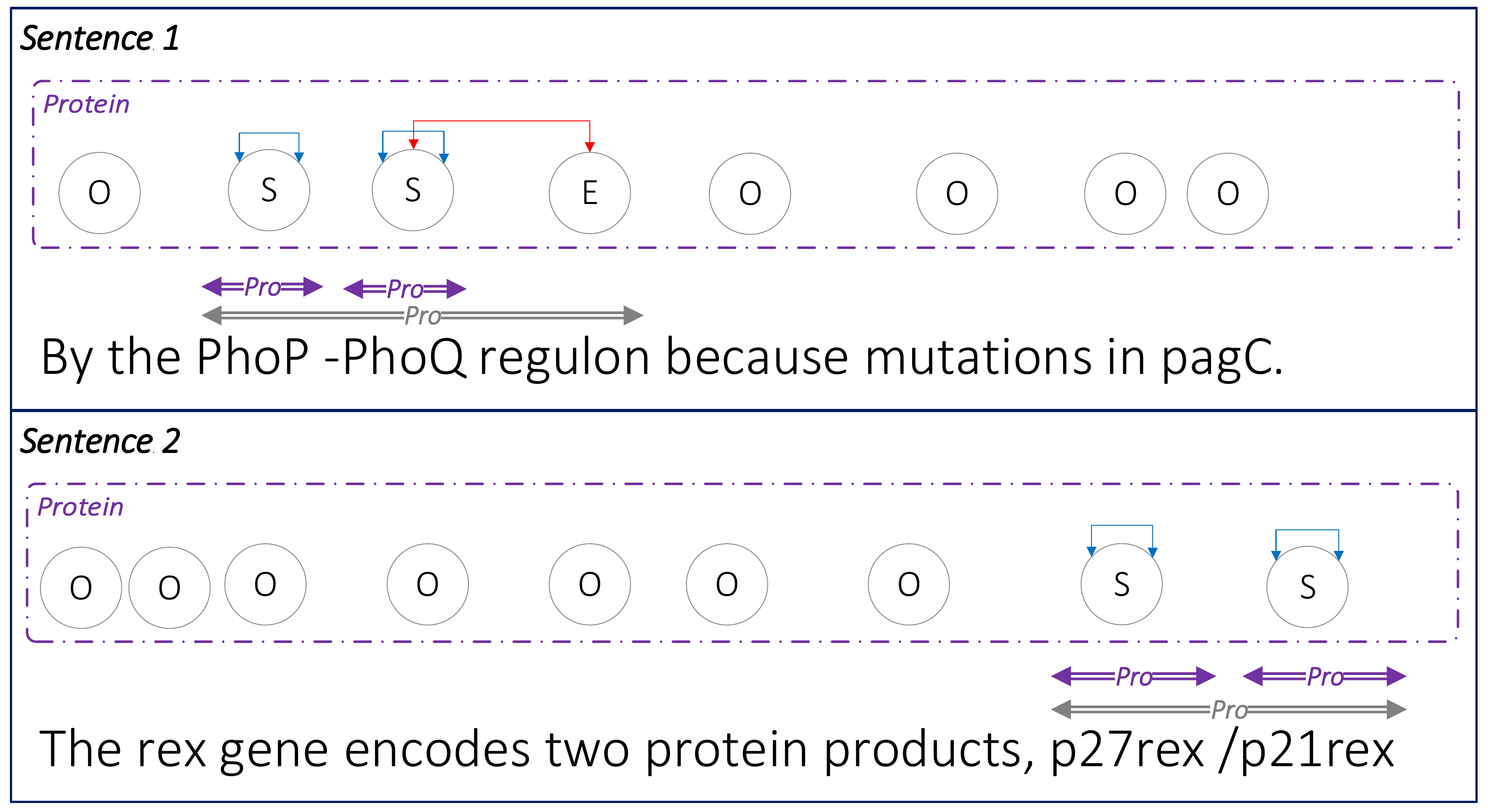}
	\caption{Example of failure tagging. The gray horizontal arrows indicate unrecognized real entities. The red arrow lines indicates the entities that was incorrectly predicted.}
	\label{failure_example}
\end{figure}

We have to admit that our labeling and decoding scheme is not theoretically perfect.
We demonstrate several examples in Figure \ref{failure_example}, in which our scheme fails to make a 100% correct identification.
These errors are caused by the inevitable ambiguity of interpretation of complex nesting scenarios when only using BIOES tagging. 
Our model adopts appropriate presuppositions to remove the ambiguity, which introduces partial errors.

So, the extent to the existence of these errors should be a great concern.
Thankfully, as Table \ref{statistic} shows, there are very few sentences containing such complex nesting.
It's actually expected since this kind of complex nesting is relatively rare in the real language environment. 
Therefore, paying the relatively high cost to design complex decoding schemes for these rare cases could be unnecessary.

\begin{table}
\centering
\caption{Statistical information on GENIA datasets.}
\label{nested_statistic}
\begin{tabular}{lccc}
\hline
\multirow{2}{*}{Info.} & \multicolumn{3}{c}{GENIA} \\ \cline{2-4} 
                          & Train   & Test   & Dev    \\ \hline
Nested entities           & 8265    & 799    & 1199   \\
NDT                       & 4554    & 423    & 625    \\
NST                       & 3909    & 382    & 605    \\
Unidentifiable entities   & 31      & 37     & 38     \\
Misidentified entities    & 67      & 75     & 76     \\
Complex sentences         & 64      & 65     & 72     \\ \hline
\end{tabular}
\end{table}

\subsubsection{Case Study}
We do the case study in Table \ref{case_study} to show our model's ability to recognize nested entities. 

The first row illustrates that our model can identify nested entities of the same type with multiple entities nested inside. 
We can see that the entity at the outer layer is the “vitamin D receptor gene (VDR) gene”,  and the entities at the inner layer are “VDR” and “vitamin D receptor”. 
All entities can be accurately identified by our model.

The second and third rows demonstrate that our model can identify precisely the nested entities of different classes. 
For the second sentence, the “GM-CSF” and “GM-CSF gene” are nested between different types of entities and can be accurately identified by our model.
As for the third sentence, the “VDR” and “VDR DNA-binding mutants” are another instance.
Our model also precisely identified them.

\begin{table*}
\centering
\caption{Case study. The blue square brackets indicate the entities predicted by the model, the red square brackets indicate the real entities. The label in the lower right of the right square bracket indicates the type of entity, and the square bracket superscript indicates the level of nesting.}
\label{case_study}
\begin{tabular}{p{\textwidth}}
\hline
In as much as these features implicate enhanced calcitriol action in gut and bone, we analyzed the $\color{blue}{\text{[}^2}$  $\color{red}{\text{[}^2}$ $\color{blue}{\text{[}^1}$ $\color{red}{\text{[}^1}$vitamin D receptor  $\color{blue}{\text{]}^1_{PRO}}$ $\color{red}{\text{]}^1_{PRO}}$ ($\color{blue}{\text{[}^1}$ $\color{red}{\text{[}^1}$VDR$\color{red}{\text{]}^1_{PRO}}$ $\color{blue}{\text{]}^1_{PRO}}$) gene$\color{red}{\text{]}^2_{PRO}}$ $\color{blue}{\text{]}^2_{PRO}}$ to ascertain whether an abnormality of this gene marks patients with intestinal hyperabsorption of calcium. \\ \hline
Our data provide strong evidence that the expression of the $\color{blue}{\text{[}^2}$  $\color{red}{\text{[}^2}$ $\color{blue}{\text{[}^1}$  $\color{red}{\text{[}^1}$ GM-CSF $\color{blue}{\text{]}^1_{PRO}}$ $\color{red}{\text{]}^1_{PRO}}$ gene $\color{blue}{\text{]}^2_{DNA}}$ $\color{red}{\text{]}^2_{DNA}}$ following T-cell activation is controlled by binding of the $\color{blue}{\text{[}^2}$  $\color{red}{\text{[}^2}$ $\color{blue}{\text{[}^1}$  $\color{red}{\text{[}^1}$  NF-kappa B $\color{blue}{\text{]}^1_{PRO}}$ $\color{red}{\text{]}^1_{PRO}}$ transcription factor $\color{blue}{\text{]}^2_{PRO}}$ $\color{red}{\text{]}^2_{PRO}}$ to a high-affinity binding site in the  $\color{blue}{\text{[}^2}$  $\color{red}{\text{[}^2}$ $\color{blue}{\text{[}^1}$  $\color{red}{\text{[}^1}$  GM-CSF $\color{blue}{\text{]}^1_{PRO}}$ $\color{red}{\text{]}^1_{PRO}}$ promoter  $\color{blue}{\text{]}^2_{DNA}}$ $\color{red}{\text{]}^2_{DNA}}$. \\ \hline
 $\color{blue}{\text{[}^2}$  $\color{red}{\text{[}^2}$ $\color{blue}{\text{[}^1}$  $\color{red}{\text{[}^1}$ VDR $\color{blue}{\text{]}^1_{PRO}}$ $\color{red}{\text{]}^1_{PRO}}$ DNA-binding mutants $\color{blue}{\text{]}^2_{DNA}}$ $\color{red}{\text{]}^2_{DNA}}$ were unable to either bind to this element in vitro or repress in vivo; the  $\color{blue}{\text{[}^2}$  $\color{red}{\text{[}^2}$ $\color{blue}{\text{[}^1}$  $\color{red}{\text{[}^1}$ VDR $\color{blue}{\text{]}^1_{PRO}}$ $\color{red}{\text{]}^1_{PRO}}$ DNA-binding domain $\color{blue}{\text{]}^2_{DNA}}$ $\color{red}{\text{]}^2_{DNA}}$ alone, however, bound the element but also could not repress IL-2 expression. \\ \hline
\end{tabular}
\end{table*}

\section{Conclusion and future work}
In this paper, we propose a type-supervised sequence labeling method based on heterogeneous star graphs for the NER task.
Specifically, we introduce type nodes to establish a typology of star connections and subsequently perform BIOES annotation on layers of each entity category.
This extended labeling method is able to recognize the vast majority of nested entities.
One of the main advantages of our approach is that it can handle entities with multiple categories which were hardly mentioned in previous work.
Also, the model achieves a relatively low time theoretical complexity in both the attention mechanism and decoding scheme.
Moreover,  we investigate the graph attention mechanism and propose our hybrid attention. 
We experiment with our heterogeneous graph model on public datasets to show the effectiveness of the approach.
Our method reaches the state-of-the-art performance on both nested and flat datasets.
We hope that our study will enhance the understanding of t sequence labeling methods and provide examples of applying graph networks in the NLP research.
In future work, we will attempt to cope with more intractable cases of nested entities, including overlapping and deep nesting.
We will also seek to expand the types of text nodes, such as introducing part-of-speech categories as text node types, in order to improve the model performance.

\section*{CRediT authorship contribution statement}
\textbf{Xueru Wen}: Conceptualization, Methodology, Software, Validation, Investigation, Writing - original draft, Writing - review \& editing. 
\textbf{Haotian Tang}: Writing - review \& editing. 
\textbf{Changjiang Zhou}: Writing - review \& editing. 
\textbf{Luguang Liang}: Writing - review \& editing.
\textbf{Hong Qi}: Project administration, Funding acquisition.
\textbf{Yu Jiang}: Writing - review \& editing, Project administration, Funding acquisition.

\section*{Declaration of competing interest}
The authors declare that they have no known competing financial interests or personal relationships that could have appeared
to influence the work reported in this paper.

\section*{Acknowledgement}
This work was supported by the National Natural Science Foundation of China under Grant 62072211 and Grant 51939003. 

%Bibliography
\bibliographystyle{unsrt}  
\bibliography{references}  

\end{document}